%% file: main.tex
\documentclass[11pt,a4paper]{article}
\usepackage[hyperref]{naaclhlt2019}
\usepackage{times}
\usepackage{latexsym}
\usepackage{xspace}
\usepackage{multirow}
\usepackage{url}
\usepackage{booktabs}
\usepackage{tikz,tikz-qtree}
\usepackage{pgfplots}
\usepackage{amssymb}
\usepackage{xfrac}
\usepackage{graphicx}
\usepackage{tablefootnote}
\usepackage{amsmath}
\usepackage{enumitem}
\pgfplotsset{compat=1.14}

\newcommand\bert{BERT\xspace}
\newcommand{\eat}[1]{\ignorespaces}

\definecolor{g-red}{HTML}{DB4437}
\definecolor{g-blue}{HTML}{4285F4}
\definecolor{g-green}{HTML}{0F9D58}
\definecolor{g-yellow}{HTML}{F4B400}
\definecolor{g-orange}{HTML}{FF9800}
\definecolor{g-grey}{HTML}{9E9E9E}

\newcommand\bertbase{BERT$_{\small \textsc{BASE}}$\xspace}
\newcommand\bertlarge{BERT$_{\small \textsc{LARGE}}$\xspace}

\title{\textbf{BERT}: Pre-training of Deep Bidirectional Transformers for \\ Language Understanding}

\author{Jacob Devlin \quad Ming-Wei Chang \quad Kenton Lee \quad Kristina Toutanova \\
  Google AI Language \\
  {\tt \{jacobdevlin,mingweichang,kentonl,kristout\}@google.com} \\}

\date{}

\aclfinalcopy 
\def\aclpaperid{1584} 

\begin{document}
\maketitle
\begin{abstract}
  \input{abstract}

\end{abstract}

\input{intro}

\input{related}

\input{bert}

\input{experiment}

\input{ablation}

\input{conclusion}

\bibliographystyle{acl_natbib}
\bibliography{lumiere}

\pagenumbering{arabic} 
\appendix                                     
\input{appendix_main}

\end{document}

%% file: abstract.tex
We introduce a new language representation model called {\bf \bert}, which stands for \textbf{B}idirectional \textbf{E}ncoder \textbf{R}epresentations from \textbf{T}ransformers. Unlike recent language representation models~\cite{peters-etal:2018:_deep, radford-etal:2018},  \bert is designed to pre-train deep bidirectional representations from unlabeled text by jointly conditioning on both left and right context in all layers. As a result, the pre-trained BERT model can be fine-tuned with just one additional output layer to create state-of-the-art models for a wide range of tasks, such as question answering and language inference, without substantial task-specific architecture modifications. 

\bert is conceptually simple and empirically powerful. It obtains new state-of-the-art results on eleven natural language processing tasks, including pushing the GLUE score to 80.5\% (7.7\% point absolute improvement), MultiNLI accuracy to 86.7\% (4.6\% absolute improvement), SQuAD v1.1 question answering Test F1 to 93.2 (1.5 point absolute improvement) and SQuAD v2.0 Test F1 to 83.1 (5.1 point absolute improvement).

%% file: intro.tex
\section{Introduction}

Language model pre-training has been shown to be effective for improving many natural language processing tasks~\cite{dai-le:2015:_semi, peters-etal:2018:_deep, radford-etal:2018, howard-ruder:2018}. These include sentence-level tasks such as natural language inference~\cite{bowman-etal:2015, williams-nangia-bowman:2018} and paraphrasing~\cite{dolan-brockett:2005:_autom}, which aim to predict the relationships between sentences by analyzing them holistically, as well as token-level tasks such as named entity recognition and question answering, where models are required to produce fine-grained output at the token level~\cite{tjong-de:2003, rajpurkar-etal:2016:_squad}.

There are two existing strategies for applying pre-trained language representations to downstream tasks: {\em feature-based} and {\em fine-tuning}. The feature-based approach, such as ELMo~\cite{peters-etal:2018:_deep}, uses task-specific architectures that include the pre-trained representations as additional features. The fine-tuning approach, such as the Generative Pre-trained Transformer (OpenAI GPT)~\cite{radford-etal:2018}, introduces minimal task-specific parameters, and is trained on the downstream tasks by simply fine-tuning {\em all} pre-trained parameters. The two approaches share the same objective function during pre-training, where they use unidirectional language models to learn general language representations.

We argue that current techniques restrict the power of the pre-trained representations, especially for the fine-tuning approaches. The major limitation is that standard language models are unidirectional, and this limits the choice of architectures that can be used during pre-training. For example, in OpenAI GPT, the authors use a left-to-right architecture, where every token can only attend to previous tokens in the self-attention layers of the Transformer~\cite{vaswani-etal:2017:_atten}. Such restrictions are sub-optimal for sentence-level tasks, and could be very harmful when applying fine-tuning based approaches to token-level tasks such as  question answering, where it is crucial to incorporate context from both directions.

In this paper, we improve the fine-tuning based approaches by proposing \bert: \textbf{B}idirectional \textbf{E}ncoder \textbf{R}epresentations from \textbf{T}ransformers. \bert alleviates the previously mentioned unidirectionality constraint by using a ``masked language model''~(MLM) pre-training objective, inspired by the Cloze task~\cite{taylor:1953:_cloze}. The masked language model randomly masks some of the tokens from the input, and the objective is to predict the original vocabulary id of the masked word based only on its context. Unlike left-to-right language model pre-training, the MLM objective enables the representation to fuse the left and the right context, which allows us to pre-train a deep bidirectional Transformer. In addition to the masked language model, we also use a ``next sentence prediction'' task that jointly pre-trains text-pair representations. The contributions of our paper are as follows:
\begin{itemize}[leftmargin=1em]
  \item We demonstrate the importance of bidirectional pre-training for language representations. Unlike \citet{radford-etal:2018}, which uses unidirectional language models for pre-training, \bert uses masked language models to enable pre-trained deep bidirectional representations. This is also in contrast to \citet{peters-etal:2018:_deep}, which uses a shallow concatenation of independently trained left-to-right and right-to-left LMs.
  \item We show that pre-trained representations reduce the need for many heavily-engineered task-specific architectures. \bert is the first fine-tuning based representation model that achieves state-of-the-art performance on a large suite of sentence-level {\em and} token-level tasks, outperforming many task-specific architectures.
  \item \bert advances the state of the art for eleven NLP tasks. 
    The code and pre-trained models are available at \url{https://github.com/google-research/bert}.
\end{itemize}

%% file: related.tex

\section{Related Work}
There is a long history of pre-training general language representations, and we briefly review the most widely-used approaches in this section.

\subsection{Unsupervised Feature-based Approaches}
Learning widely applicable representations of words has been an active area of research for decades, including non-neural~\cite{brown-etal:1992:_class, ando-zhang:2005, blitzer-mcdonald-pereira:2006:_domain} and neural~\cite{mikolov-etal:2013, pennington-socher-manning:2014:_glove} methods. Pre-trained word embeddings are an integral part of modern NLP systems, offering significant improvements over embeddings learned from scratch~\cite{turian-ratinov-bengio:2010:_word_repres}. To pre-train word embedding vectors, left-to-right language modeling objectives have been used~\cite{minh09}, as well as objectives to  discriminate correct from incorrect words in left and right context~\cite{mikolov-etal:2013}.

These approaches have been generalized to coarser granularities, such as sentence embeddings~\cite{kiros-etal:2015:_skip, logeswaran2018an} or paragraph embeddings~\cite{le-mikolov:2014:_distr}. To train sentence representations, prior work has used objectives to rank candidate next sentences  \cite{DBLP:journals/corr/JerniteBS17, logeswaran2018an},  left-to-right generation of next sentence words given a representation of the previous sentence~\cite{kiros-etal:2015:_skip}, or denoising auto-encoder derived objectives~\cite{hill16}.

\begin{figure*}[t!]
\begin{center}
\includegraphics[width=1\textwidth]{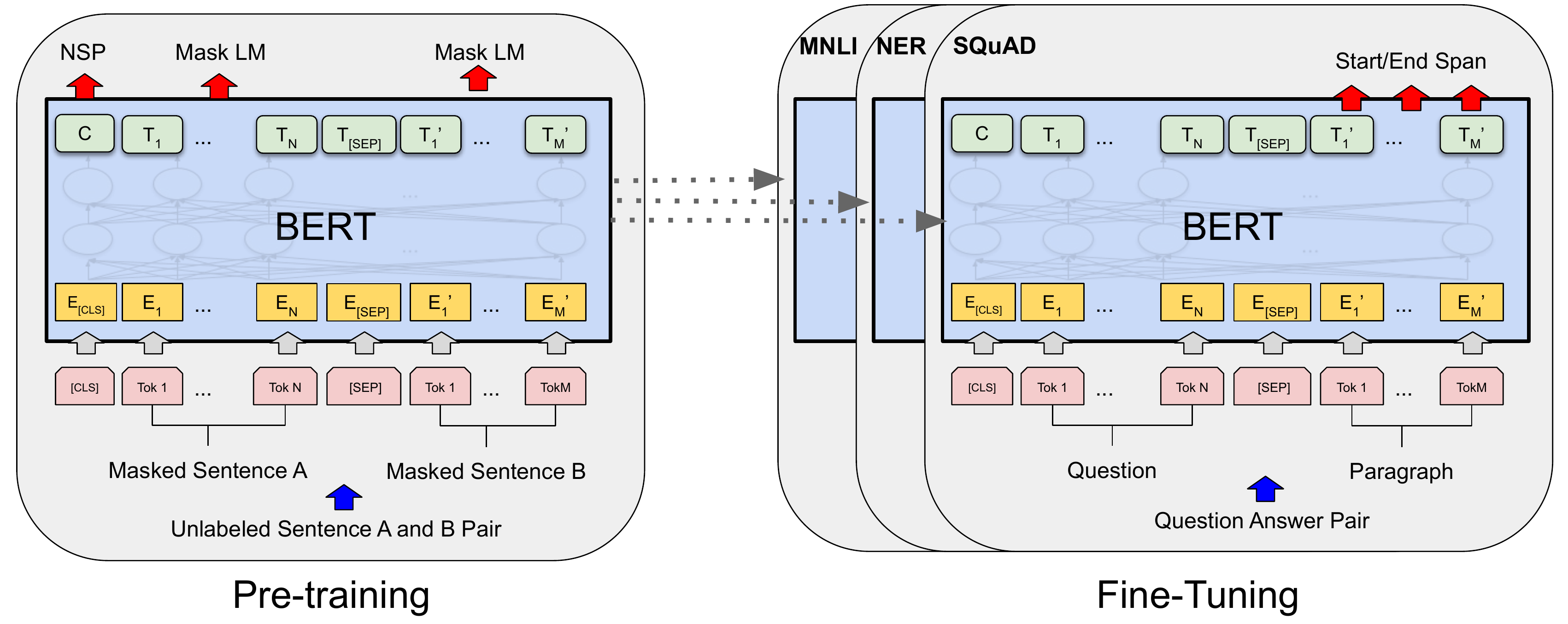}
\end{center}
\caption{Overall pre-training and fine-tuning procedures for BERT. Apart from output layers, the same architectures are used in both pre-training and fine-tuning. The same pre-trained model parameters are used to initialize models for different down-stream tasks.  During fine-tuning, all parameters are fine-tuned. {\tt [CLS]} is a special symbol added
in front of every input example, and {\tt [SEP]} is a special separator token (e.g. separating  questions/answers).}
\label{fig:bert_overall}
\end{figure*}

ELMo and its predecessor~\cite{peters-etal:2017:_semi, peters-etal:2018:_deep} generalize traditional word embedding research along a different dimension. They extract \emph{context-sensitive} features from a left-to-right and a right-to-left language model. The contextual representation of each token is the concatenation of the left-to-right and right-to-left representations. When integrating contextual word embeddings with existing task-specific architectures, ELMo advances the state of the art for several major NLP benchmarks~\cite{peters-etal:2018:_deep} including question answering~\cite{rajpurkar-etal:2016:_squad}, sentiment analysis~\cite{socher-etal:2013:_recur}, and named entity recognition~\cite{tjong-de:2003}.
\citet{melamud2016context2vec} proposed learning contextual representations through a task to predict a single word from both left and right context
using LSTMs. Similar to ELMo, their model is feature-based and not deeply bidirectional. 
\citet{fedus2018maskgan} shows that the cloze task can be used to improve the robustness of text generation models.

\subsection{Unsupervised Fine-tuning Approaches}

As with the feature-based approaches, the first works in this direction only pre-trained word embedding parameters from unlabeled text ~\cite{collobert-weston:2008}.  

More recently, sentence or document encoders which produce contextual token representations have been pre-trained from unlabeled text and fine-tuned for a supervised downstream task~\cite{dai-le:2015:_semi, howard-ruder:2018, radford-etal:2018}.  The advantage of these approaches is that few parameters need to be learned from scratch. At least partly due to this advantage, OpenAI GPT~\cite{radford-etal:2018} achieved previously state-of-the-art results on many sentence-level tasks from the GLUE benchmark~\cite{wang-etal:2018:_glue}.  Left-to-right language modeling and auto-encoder objectives have been used for pre-training such models~\cite{howard-ruder:2018, radford-etal:2018,dai-le:2015:_semi}.

\subsection{Transfer Learning from Supervised Data}

There has also been work showing effective transfer from supervised tasks with large datasets, such as natural language inference~\cite{conneau-EtAl:2017:EMNLP2017} and machine translation~\cite{mccann-etal:2017:_learn_trans}. 
Computer vision research has also demonstrated the importance of transfer learning from large pre-trained models, where an effective recipe is to fine-tune models pre-trained with ImageNet~\cite{imagenet_cvpr09, yosinski2014transferable}.

%% file: bert.tex

\section{BERT}
\label{sec:bert}

We introduce BERT and its detailed implementation in this section. There are two steps in our framework: {\em pre-training} and {\em fine-tuning}. 
During pre-training, the model is trained on unlabeled data over different pre-training tasks.
For fine-tuning, the BERT model is first initialized with the pre-trained parameters, and all of the parameters are fine-tuned using labeled data from the downstream tasks. 
Each downstream task has separate fine-tuned models, even though they
are initialized with the same pre-trained parameters. The question-answering example in Figure~\ref{fig:bert_overall} will serve as a running example for this section.

A distinctive feature of BERT is its unified architecture across different tasks.
There is minimal difference between the pre-trained architecture and the final downstream architecture.

\paragraph{Model Architecture}
BERT's model architecture is a multi-layer bidirectional Transformer encoder based on the original implementation described in \citet{vaswani-etal:2017:_atten} and released in the {\tt tensor2tensor} library.\footnote{https://github.com/tensorflow/tensor2tensor} Because the use of Transformers has become common and our implementation is almost identical to the original, we will omit an exhaustive background description of the model architecture and refer readers to \citet{vaswani-etal:2017:_atten} as well as excellent guides such as ``The Annotated Transformer.''\footnote{http://nlp.seas.harvard.edu/2018/04/03/attention.html}

In this work, we denote the number of layers (i.e., Transformer blocks) as $L$, the hidden size as $H$, and the number of self-attention heads as $A$.\footnote{In all cases we set the feed-forward/filter size to be $4H$, i.e., 3072 for the $H=768$ and 4096 for the $H=1024$.} We primarily report results on two model sizes: {\bf \bertbase} (L=12, H=768, A=12, Total Parameters=110M) and {\bf \bertlarge} (L=24, H=1024, A=16, Total Parameters=340M).

\bertbase was chosen to have the same model size as OpenAI GPT for comparison purposes. Critically, however, the BERT Transformer uses bidirectional self-attention, while the GPT Transformer uses constrained self-attention where every token can only attend to context to its left.\footnote{We note that in the literature the bidirectional Transformer is often referred to as a ``Transformer encoder'' while the left-context-only version is referred to as a ``Transformer decoder'' since it can be used for text generation.}

\paragraph{Input/Output Representations} 
To make BERT handle a variety of down-stream tasks,
our input representation is able to unambiguously represent both a single sentence and a pair of  sentences (e.g., $\langle$ Question, Answer $\rangle$) in one token sequence. Throughout this work, a ``sentence'' can be an arbitrary span of contiguous text, rather than an actual linguistic sentence. A ``sequence'' refers to the input token sequence to BERT, which may be a single sentence or two sentences packed together.

We use WordPiece embeddings \cite{wu-etal:2016:_googl} with a 30,000 token vocabulary.
The first token of every sequence is always a special classification token ({\tt [CLS]}). The final hidden state corresponding to this token is used as the aggregate sequence representation for classification tasks. 
Sentence pairs are packed together into a single sequence. We differentiate the sentences in two ways. First, we separate them with a special token ({\tt [SEP]}). Second, we add a learned embedding to every token indicating whether it belongs to sentence {\tt A} or sentence {\tt B}. 
As shown in Figure~\ref{fig:bert_overall}, we denote 
input embedding as $E$, the final hidden vector of the special {\tt [CLS]} token as $C \in \mathbb{R}^{H}$, and the final hidden vector for the $i^{\rm th}$ input token as $T_i \in \mathbb{R}^H$.

\begin{figure*}[ht]
\begin{center}
\hspace{-0.2in}
\includegraphics[width=360px]{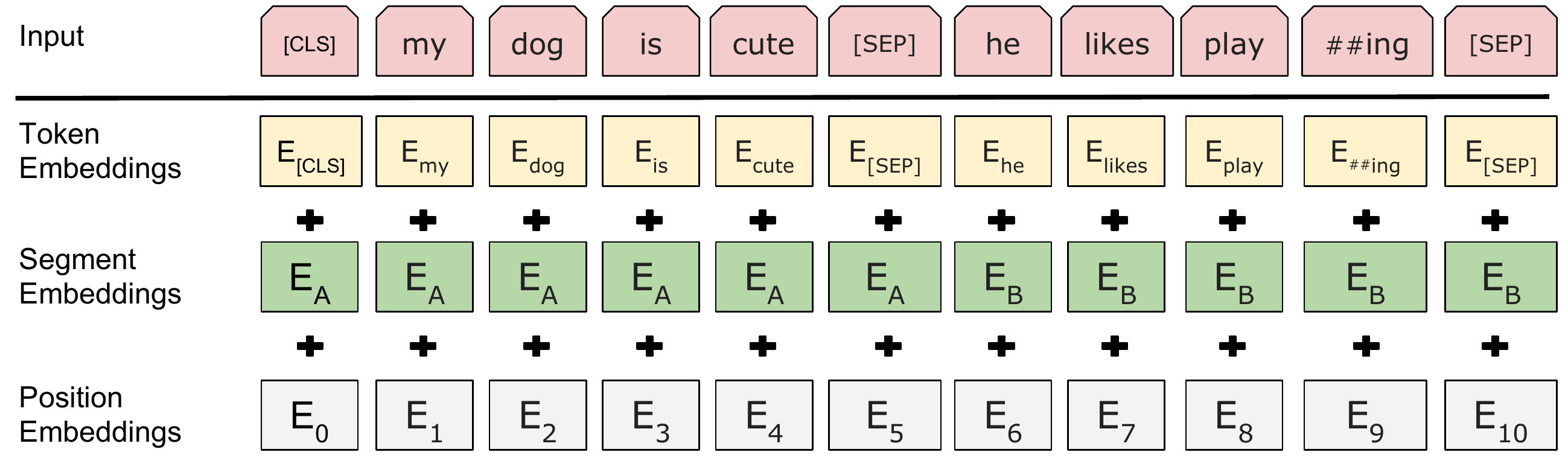}
\end{center}
\caption{BERT input representation. The input embeddings are the sum of the token embeddings, the segmentation embeddings and the position embeddings.}
\label{fig:input_embeddings}
\end{figure*}

For a given token, its input representation is constructed by summing the corresponding token, segment, and position embeddings. 
A visualization of this construction can be seen in Figure~\ref{fig:input_embeddings}.

\subsection{Pre-training BERT}
\label{sec:pretraining_tasks}

Unlike \citet{peters-etal:2018:_deep} and \citet{radford-etal:2018}, we do not use traditional left-to-right or right-to-left language models to pre-train BERT. Instead, we pre-train BERT using two unsupervised tasks, described in this section. This step
is presented in the left part of Figure~\ref{fig:bert_overall}.

\paragraph{Task \#1: Masked LM}
Intuitively, it is reasonable to believe that a deep bidirectional model is strictly more powerful than either a left-to-right model or the shallow concatenation of a left-to-right and a right-to-left model. Unfortunately, standard conditional language models can only be trained left-to-right {\it or} right-to-left, since bidirectional conditioning would 
allow each word to indirectly ``see itself'', and
the model could trivially predict the target word in a multi-layered context.

In order to train a deep bidirectional representation, we simply mask some percentage of the input tokens at random, and then predict those masked tokens. We refer to this procedure as a ``masked LM'' (MLM), although it is often referred to as a {\it Cloze} task in the literature~\cite{taylor:1953:_cloze}. In this case, the final hidden vectors corresponding to the mask tokens are fed into an output softmax over the vocabulary, as in a standard LM. In all of our experiments, we mask 15\% of all WordPiece tokens in each sequence at random. In contrast to denoising auto-encoders \cite{vincent:2008}, we only predict the masked words rather than reconstructing the entire input.  

Although this allows us to obtain a bidirectional pre-trained model, a downside is that we are creating a mismatch between pre-training and fine-tuning, since the {\tt [MASK]} token does not appear during fine-tuning. To mitigate this, we do not always replace ``masked'' words with the actual {\tt [MASK]} token. The training data generator chooses 15\% of the token positions at random for prediction. If the $i$-th token is chosen, we replace the $i$-th token with (1) the {\tt [MASK]} token 80\% of the time (2) a random token 10\% of the time (3) the unchanged $i$-th token 10\% of the time. Then, $T_i$ will be used to predict the original token with cross entropy loss.
We compare variations of this procedure in Appendix~\ref{appendix:sec:different_masks}.
\vspace{.2cm}

\paragraph{Task \#2: Next Sentence Prediction (NSP)}
Many important downstream tasks such as Question Answering (QA) and Natural Language Inference (NLI) are based on understanding the {\it relationship} between two sentences, which is not directly captured by language modeling. In order to train a model that understands sentence relationships, we pre-train for a binarized {\it next sentence prediction} task that can be trivially generated from any monolingual corpus. Specifically, when choosing the sentences {\tt A} and {\tt B} for each pre-training example, 50\% of the time {\tt B} is the actual next sentence that follows {\tt A} (labeled
 as {\tt {\small IsNext}}), and 50\% of the time it is a random sentence from the corpus
 (labeled as {\tt {\small NotNext}}). 
As we show in Figure~\ref{fig:bert_overall}, $C$ is used for  next sentence prediction (NSP).\footnote{The final model achieves 97\%-98\% accuracy on NSP.} Despite its simplicity, we demonstrate in Section~\ref{sec:task_ablation} that pre-training towards this task is very beneficial to both QA and NLI.
\footnote{The vector $C$ is not a meaningful sentence representation without fine-tuning, since it was trained with NSP.}
The NSP task is closely related to representation-learning objectives used in \citet{DBLP:journals/corr/JerniteBS17} and \citet{logeswaran2018an}. However, in prior work, only sentence embeddings are transferred to down-stream tasks, where BERT transfers all parameters to initialize end-task model parameters.

\vspace{.3cm}
\noindent\textbf{Pre-training data}
The pre-training procedure largely follows the existing literature on language model pre-training. For the pre-training corpus we use the BooksCorpus (800M words)~\cite{zhu:2015} and English Wikipedia (2,500M words). For Wikipedia we extract only the text passages and ignore lists, tables, and headers. It is critical to use a document-level corpus rather than a shuffled sentence-level corpus such as the Billion Word Benchmark \cite{chelba-etal:2013:_one} in order to extract long contiguous sequences.

\subsection{Fine-tuning BERT}
\label{sec:finetuning_procedure}

Fine-tuning is straightforward since the self-attention mechanism in the Transformer allows BERT to model many downstream tasks---whether they involve single text or text pairs---by swapping out the appropriate inputs and outputs.
For applications involving text pairs, a common pattern is to independently encode text pairs before applying bidirectional cross attention, such as~\newcite{parikh-etal:2016, bidaf}. BERT instead uses the self-attention mechanism to unify these two stages, as encoding 
a concatenated text pair with self-attention effectively includes \emph{bidirectional} cross attention between two sentences.

For each task, we simply plug in the task-specific inputs and outputs into BERT and fine-tune all the parameters end-to-end. 
At the input, sentence {\tt A} and sentence {\tt B} from pre-training are analogous to (1) sentence pairs in paraphrasing, (2) hypothesis-premise pairs in entailment, (3) question-passage pairs in question answering, and (4) a degenerate text-$\varnothing$ pair in text classification or sequence tagging. At the output, the token representations are fed into an output layer for token-level tasks, such as sequence tagging or question answering, and the {\tt [CLS]} representation is fed into an output layer for classification, such as entailment or sentiment analysis.

Compared to pre-training, fine-tuning is relatively inexpensive. All of the results in the paper can be replicated in at most 1 hour on a single Cloud TPU, or a few hours on a GPU, starting from the exact same pre-trained model.\footnote{For example, the BERT SQuAD model can be trained in around 30 minutes on a single Cloud TPU to achieve a Dev F1 score of 91.0\%.}
We describe the task-specific details in the corresponding subsections of Section~\ref{sec:experiments}. 
More details can be found in Appendix~\ref{appendix:sec:fine_tune_details_and_figures}.

%% file: experiment.tex
\input{glue_official_tab}
\section{Experiments}
\label{sec:experiments}

In this section, we present BERT fine-tuning results on 11 NLP tasks. 

\subsection{GLUE}
\label{sec:glue}

The General Language Understanding Evaluation (GLUE) benchmark~\cite{wang-etal:2018:_glue} is a collection of diverse natural language understanding tasks.
Detailed descriptions of GLUE datasets are included in Appendix~\ref{appendix:sec:glue}.

To fine-tune on GLUE, we represent the input sequence (for single sentence or sentence pairs) as described in Section~\ref{sec:bert}, and use the final hidden vector $C \in \mathbb{R}^{H}$ corresponding to the first input token ({\tt [CLS]}) as the aggregate representation.
The only new parameters introduced during fine-tuning are  classification layer weights $W \in \mathbb{R}^{K \times H}$, where $K$ is the number of labels. We compute a standard classification loss with $C$ and $W$, i.e., $\log({\rm softmax}(CW^T))$.

We use a batch size of 32 and fine-tune for 3 epochs over the data for all GLUE tasks. For each task, we selected the best fine-tuning learning rate (among 5e-5, 4e-5, 3e-5, and 2e-5) on the Dev set. Additionally, for \bertlarge we found that fine-tuning was sometimes unstable on small datasets, so we ran several random restarts and selected the best model on the Dev set. With random restarts, we use the same pre-trained checkpoint but perform different fine-tuning data shuffling and classifier layer initialization.\footnote{The GLUE data set distribution does not include the Test labels, and we only made a single GLUE evaluation server submission for each of \bertbase and \bertlarge.}

Results are presented in Table~\ref{tab:glue_official}. Both \bertbase and \bertlarge outperform all systems on all tasks by a substantial margin, obtaining 4.5\% and 7.0\% respective average accuracy improvement over the prior state of the art. Note that \bertbase and OpenAI GPT are nearly identical in terms of model architecture apart from the attention masking. For the largest and most widely reported GLUE task, MNLI, BERT obtains a 4.6\% absolute accuracy improvement. On the official GLUE leaderboard\footnote{https://gluebenchmark.com/leaderboard},
\bertlarge obtains a score of 80.5, compared to OpenAI GPT, which obtains 72.8 as of the date of writing.

We find that \bertlarge significantly outperforms \bertbase across all tasks, especially those with very little training data. The effect of model size is explored more thoroughly in Section~\ref{sec:model_size_ablation}.

\subsection{SQuAD v1.1}
\label{sec:squad}

The Stanford Question Answering Dataset (SQuAD v1.1) is a collection of 100k crowdsourced question/answer pairs~\cite{rajpurkar-etal:2016:_squad}. Given a question and a passage from Wikipedia containing the answer, the task is to predict the answer text span in the passage. 

\eat{
For example: 

\begin{itemize}
\item Input Question: \\{\tt {\scriptsize  Where do water droplets collide with ice crystals to form precipitation?}}
\item Input Paragraph: \\{\tt {\scriptsize  ... Precipitation forms as smaller droplets coalesce via collision with other rain drops or ice crystals within a cloud. ...}}
\item Output Answer: \\ {\tt {\scriptsize  within a cloud}}
\end{itemize}
}

As shown in Figure~\ref{fig:bert_overall}, in the question answering task,
we represent the input question and passage as a single packed sequence, with the question using the {\tt A} embedding and the passage using the {\tt B} embedding. We only introduce a start vector $S \in \mathbb{R}^H$ and an end vector $E \in \mathbb{R}^H$ during fine-tuning.
The probability of word $i$ being the start of the answer span is computed as a dot product between $T_i$ and $S$ followed by a softmax over all of the words in the paragraph: $P_i = \frac{e^{S{\cdot}T_i}}{\sum_j e^{S{\cdot}T_j}}$. The analogous formula is used for the end of the answer span. The score of a candidate span from position $i$ to position $j$ is defined as  $S{\cdot}T_i + E{\cdot}T_j$, and the maximum scoring span where $j \geq i$ is used as a prediction. The training objective is the sum of the log-likelihoods of the correct start and end positions. We fine-tune for 3 epochs with a learning rate of 5e-5 and a batch size of 32. 

\input{squad_tab}

Table~\ref{tab:squad_results} shows top leaderboard entries as well as results from top published systems~\cite{bidaf,clark-gardner:2018:_simpl,peters-etal:2018:_deep,hu2017reinforced}.
The top results from the SQuAD leaderboard do not have up-to-date public system descriptions available,\footnote{QANet is described in \newcite{yu-etal:2018:_qanet}, but the system has improved substantially after publication.} and are allowed to use any public data when training their systems. We therefore use modest data augmentation in our system by first fine-tuning on TriviaQA~\cite{joshi-etal:2017:_triviaq} befor fine-tuning on SQuAD.

Our best performing system outperforms the top leaderboard system by +1.5 F1 in ensembling and +1.3 F1 as a single system. In fact, our single BERT model outperforms the top ensemble system in terms of F1 score. Without TriviaQA fine-tuning data, we only lose 0.1-0.4 F1, still outperforming all existing systems by a wide margin.\footnote{The TriviaQA data we used consists of paragraphs from TriviaQA-Wiki formed of the first 400 tokens in documents, that contain at least one of the provided possible answers.}

\subsection{SQuAD v2.0}

The SQuAD 2.0 task extends the SQuAD 1.1 problem definition by allowing for the possibility that no short answer exists in the provided paragraph, making the problem more realistic.

We use a simple approach to extend the SQuAD v1.1 BERT model for this task. We treat questions that do not have an answer as having an answer span with start and end at the {\tt [CLS]} token. The probability space for the start and end answer span positions is extended to include the position of the {\tt [CLS]} token. For prediction, we compare the score of the no-answer span: $s_{\tt null} = S{\cdot}C + E{\cdot}C$ to the score of the best non-null span $\hat{s_{i,j}}$ =  $ {\tt max}_{j \geq i} S{\cdot}T_i + E{\cdot}T_j$. We predict a non-null answer when  $\hat{s_{i,j}} > s_{\tt null} + \tau $, where the threshold $\tau$ is selected on the dev set to maximize F1. We did not use TriviaQA data for this model. We fine-tuned for 2 epochs with a learning rate of 5e-5 and a batch size of 48.

The results compared to prior leaderboard entries and top published work \cite{unet,slqa} are shown in Table~\ref{tab:squad2_results}, excluding systems that use BERT as one of their components. We observe a +5.1 F1 improvement over the previous best system.

\subsection{SWAG}
\label{sec:swag}
The Situations With Adversarial Generations (SWAG) dataset contains 113k sentence-pair completion examples that evaluate grounded commonsense inference~\cite{zellers2018swag}. Given a sentence, the task is to choose the most plausible continuation among four choices.

When fine-tuning on the SWAG dataset, we construct four input sequences, each containing the concatenation of the given sentence (sentence {\tt A}) and a possible continuation (sentence {\tt B}). The only task-specific parameters introduced is a vector whose dot product with the {\tt [CLS]} token representation $C$ denotes a score for each choice which is normalized with a softmax layer.

We fine-tune the model for 3 epochs with a learning rate of 2e-5 and a batch size of 16. Results are presented in Table~\ref{tab:swag_official}. \bertlarge outperforms the authors' baseline ESIM+ELMo system by +27.1\% and OpenAI GPT by 8.3\%.

\input{swag_official_tab}

%% file: glue_official_tab.tex
\begin{table*}[t]
\small
\renewcommand{\arraystretch}{1.2}
\begin{center}
 \begin{tabular*}{\textwidth}{l@{\extracolsep{\fill}}cccccccc c}
    \toprule
System             &  MNLI-(m/mm)    & QQP        & QNLI       & SST-2      & CoLA       & STS-B      & MRPC       & RTE        & {\bf Average} \\
                  & 392k            & 363k       & 108k       & 67k        & 8.5k       & 5.7k       & 3.5k       & 2.5k       & -          \\ 
\hline
Pre-OpenAI SOTA    & 80.6/80.1       & 66.1       & 82.3       & 93.2       & 35.0       & 81.0       & 86.0       & 61.7       & 74.0       \\
BiLSTM+ELMo+Attn   & 76.4/76.1       & 64.8       & 79.8       & 90.4       & 36.0       & 73.3       & 84.9       & 56.8       & 71.0       \\
OpenAI GPT         & 82.1/81.4       & 70.3       & 87.4       & 91.3       & 45.4       & 80.0       & 82.3       & 56.0       & 75.1       \\

\hline
\bertbase          & 84.6/83.4       & 71.2       & 90.5       & 93.5       & 52.1       & 85.8       & 88.9       & 66.4       & 79.6       \\
\bertlarge         & {\bf 86.7/85.9} & {\bf 72.1} & {\bf 92.7} & {\bf 94.9} & {\bf 60.5} & {\bf 86.5} & {\bf 89.3} & {\bf 70.1} & {\bf 82.1} \\
    \bottomrule
   \end{tabular*}
   \caption{GLUE Test results, scored by the evaluation server ({\small \url{https://gluebenchmark.com/leaderboard}}). The number below each task denotes the number of training examples. The ``Average'' column is slightly different than the official GLUE score, since we exclude the problematic WNLI set.\footnote{See question 10 in \url{https://gluebenchmark.com/faq}.} 
   BERT and OpenAI GPT are single-model, single task. F1 scores are reported for QQP and MRPC, Spearman correlations are reported for STS-B, and accuracy scores are reported for
   the other tasks. We exclude entries that use BERT as one of their components.}
   \label{tab:glue_official}
\end{center}
\end{table*}
\footnotetext{See (10) in \url{https://gluebenchmark.com/faq}.}

%% file: squad_tab.tex

\begin{table}[t]
\begin{center}
{\small
\begin{tabular}{@{}lcccc@{}}
  \toprule
  \multicolumn{1}{c}{System} & \multicolumn{2}{c}{Dev} & \multicolumn{2}{c}{Test} \\
  & EM & F1 & EM & F1 \\
  \midrule

  \multicolumn{5}{c}{Top Leaderboard Systems (Dec 10th, 2018)} \\
  Human                & - & - & 82.3 & 91.2 \\ 
  \#1 Ensemble - nlnet & - & - & 86.0 & 91.7 \\ 
  \#2 Ensemble - QANet & - & - & 84.5 & 90.5 \\ 
  \midrule
  \multicolumn{5}{c}{Published}     \\
 BiDAF+ELMo (Single)    & -    & 85.6  & -     & 85.8    \\ 
 R.M. Reader (Ensemble) & 81.2 & 87.9  & 82.3  & 88.5 \\ 
  \midrule
  \multicolumn{5}{c}{Ours} \\
  \bertbase (Single)     & 80.8 & 88.5 & -   & - \\ 
  \bertlarge (Single)    & 84.1 & 90.9 & -   & - \\ 
  \bertlarge (Ensemble)  & 85.8 & 91.8 & -   & - \\ 
  \bertlarge (Sgl.+TriviaQA) & {\bf 84.2} & {\bf 91.1} & {\bf 85.1} & {\bf 91.8} \\ 
  \bertlarge (Ens.+TriviaQA) & {\bf 86.2} & {\bf 92.2} & {\bf 87.4} & {\bf 93.2}
 \\ 

\bottomrule
\end{tabular}
} 
\end{center}
\caption{\label{tab:squad_results} SQuAD 1.1 results. The BERT ensemble is 7x systems which use different pre-training checkpoints and fine-tuning seeds.}
\end{table}

\begin{table}[t]
\begin{center}
{\small
\begin{tabular}{@{}lcccc@{}}
  \toprule
  \multicolumn{1}{c}{System} & \multicolumn{2}{c}{Dev} & \multicolumn{2}{c}{Test} \\
  & EM & F1 & EM & F1 \\
  \midrule

\multicolumn{5}{c}{Top Leaderboard Systems (Dec 10th, 2018)} \\
  Human                & 86.3 & 89.0 & 86.9& 89.5 \\ 
  \#1 Single - MIR-MRC (F-Net)   & - & - & 74.8 & 78.0 \\ 
  \#2 Single - nlnet   & - & - & 74.2 & 77.1 \\ 
  \midrule
  \multicolumn{5}{c}{Published}     \\
 unet (Ensemble) & - & -  & 71.4  & 74.9 \\ 
 SLQA+ (Single) & - & & 71.4 & 74.4\\
  \midrule
  \multicolumn{5}{c}{Ours} \\
  \bertlarge (Single)    & 78.7 & 81.9 & 80.0   &  83.1
 \\ 

\bottomrule
\end{tabular}
} 
\end{center}
\caption{\label{tab:squad2_results} SQuAD 2.0 results. We exclude entries that use BERT as one of their components.}
\end{table}

%% file: swag_official_tab.tex
\begin{table}[tb]
\begin{center}
{
\small
 \begin{tabular}{lcc}
    \toprule
System             &  Dev    & Test\\ 
\midrule
ESIM+GloVe   & 51.9       & 52.7 \\
ESIM+ELMo    & 59.1       & 59.2 \\
OpenAI GPT    & -       & 78.0 \\
\midrule
\bertbase    & 81.6       &  - \\
\bertlarge   & {\bf 86.6} & {\bf 86.3} \\
\midrule
Human (expert)$^\dagger$ & - & 85.0 \\
Human (5 annotations)$^\dagger$        & -          & 88.0 \\
\bottomrule
\end{tabular}
}
\caption{SWAG Dev and Test accuracies. 
$^\dagger$Human performance is measured with 100 samples, as reported in the SWAG paper.}
\label{tab:swag_official}
\end{center}
\end{table}

%% file: ablation.tex
\section{Ablation Studies}
\label{sec:ablation}
In this section, we perform ablation experiments over a number of facets of BERT in order to better understand their relative importance. Additional ablation studies can be found in Appendix~\ref{appendix:sec:more_ablation_studies}.

\subsection{Effect of Pre-training Tasks}

\label{sec:task_ablation}
We demonstrate the importance of the deep bidirectionality of BERT by evaluating two pre-training objectives using exactly the same pre-training data, fine-tuning scheme, and hyperparameters as \bertbase:
\vspace{0.3cm}
\\
\noindent\textbf{No NSP}: A bidirectional model which is trained using the ``masked LM'' (MLM) but without the ``next sentence prediction'' (NSP) task.\\
\noindent\textbf{LTR \& No NSP}: A left-context-only model which is trained using a standard Left-to-Right (LTR) LM, rather than an MLM. The left-only constraint was also applied at fine-tuning, because  removing it introduced a pre-train/fine-tune mismatch that degraded downstream performance. Additionally, this model was pre-trained without the NSP task. This is directly comparable to OpenAI GPT, but using our larger training dataset, our input representation, and our fine-tuning scheme.
\input{direction_ablation_tab}

We first examine the impact brought by the NSP task. In Table~\ref{tab:task_ablation}, we show that removing NSP hurts performance significantly on QNLI, MNLI, and SQuAD 1.1. Next, we evaluate the impact of training bidirectional representations by comparing ``No NSP'' to ``LTR \& No NSP''. The LTR model performs worse than the MLM model on all tasks, with large drops on MRPC and SQuAD.

For SQuAD it is intuitively clear that a LTR model will perform poorly at token predictions, since the token-level hidden states have no right-side context.
In order to make a good faith attempt at strengthening the LTR system, we added a randomly initialized BiLSTM on top. This does significantly improve results on SQuAD, but the results are still far worse than those of the pre-trained bidirectional models. The BiLSTM hurts performance on the GLUE tasks. 

We recognize that it would also be possible to train separate LTR and RTL models and represent each token as the concatenation of the two models, as ELMo does. However: (a) this is twice as expensive as a single bidirectional model; (b) this is non-intuitive for tasks like QA, since the RTL model would not be able to condition the answer on the question; (c) this it is strictly less powerful than a deep bidirectional model, since it can use both left and right context at every layer.

\subsection{Effect of Model Size}
\label{sec:model_size_ablation}

In this section, we explore the effect of model size on fine-tuning task accuracy. We trained a number of BERT models with a differing number of layers, hidden units, and attention heads, while otherwise using the same hyperparameters and training procedure as described previously.

Results on selected GLUE tasks are shown in Table~\ref{tab:size_ablation}. In this table, we report the average Dev Set accuracy from 5 random restarts of fine-tuning. We can see that larger models lead to a strict accuracy improvement across all four datasets, even for MRPC which only has 3,600 labeled training examples, and is substantially different from the pre-training tasks. It is also perhaps surprising that we are able to achieve such significant improvements on top of models which are already quite large relative to the existing literature. For example, the largest Transformer explored in \citet{vaswani-etal:2017:_atten} is (L=6, H=1024, A=16) with 100M parameters for the encoder, and the largest Transformer we have found in the literature is (L=64, H=512, A=2) with 235M parameters \cite{alrfou:2018}. By contrast, \bertbase contains 110M parameters and \bertlarge contains 340M parameters.

\input{size_ablation_tab}

It has long been known that increasing the model size will lead to continual improvements on large-scale tasks such as machine translation and language modeling, which is demonstrated by the LM perplexity of held-out training data shown in Table~\ref{tab:size_ablation}. However, we believe that this is the first work to demonstrate convincingly that scaling to extreme model sizes also leads to large improvements on very small scale tasks, provided that the model has been sufficiently pre-trained. \citet{peters2018dissecting} presented mixed results on the downstream task impact of increasing the pre-trained bi-LM size from two to four layers and \citet{melamud2016context2vec} mentioned in passing that increasing hidden dimension size from 200 to 600 helped, but increasing further to 1,000 did not bring further improvements. Both of these prior works used a feature-based approach --- we hypothesize that when the model is fine-tuned directly on the downstream tasks and uses only a very small number of randomly initialized additional parameters, the task-specific models can benefit from the larger, more expressive pre-trained representations even when downstream task data is very small.

\subsection{Feature-based Approach with BERT}
\label{sec:ner}
All of the BERT results presented so far have used the fine-tuning approach, where a simple classification layer is added to the pre-trained model, and all parameters are jointly fine-tuned on a downstream task. However, the feature-based approach, where fixed features are extracted from the pre-trained model, has certain advantages. First, not all
tasks can be easily represented by a Transformer encoder architecture, and therefore require a task-specific model architecture to be added. Second, there are major computational benefits to
pre-compute an expensive representation of the training data once and then run many experiments with 
cheaper
models on top of this representation. 

In this section, we compare the two approaches by applying BERT to the CoNLL-2003 Named Entity Recognition (NER) task~\cite{tjong-de:2003}. In the input to BERT, we use a case-preserving WordPiece model, and we include the maximal document context provided by the data. Following standard practice, we formulate this as a tagging task but do not use a CRF layer in the output. We use the representation of the first sub-token as the input to the token-level classifier over the NER label set.

To ablate the fine-tuning approach, we apply the feature-based approach by extracting the activations from one or more layers {\it without} fine-tuning any parameters of BERT. These contextual embeddings are used as input to a randomly initialized two-layer 768-dimensional BiLSTM before the classification layer.

Results are presented in Table~\ref{tab:pretrained_embeddings}. \bertlarge performs competitively with state-of-the-art methods. The best performing method concatenates the token representations from the top four hidden layers of the pre-trained Transformer, which is only 0.3 F1 behind fine-tuning the entire model. This demonstrates that BERT is effective for both fine-tuning and feature-based approaches.

\input{pretrained_embeddings_tab}

%% file: direction_ablation_tab.tex
\begin{table}[t]
\small
 \begin{tabular}{@{}lccccc@{}}
    \toprule
              & \multicolumn{5}{c}{Dev Set} \\
   Tasks & MNLI-m & QNLI & MRPC & SST-2 & SQuAD     \\
         & (Acc) & (Acc) & (Acc) & (Acc) & (F1)     \\
     \midrule
\bertbase       & 84.4 & 88.4 & 86.7 & 92.7 & 88.5 \\
No NSP          & 83.9 & 84.9 & 86.5 & 92.6 & 87.9 \\
LTR \& No NSP   & 82.1 & 84.3 & 77.5 & 92.1 & 77.8 \\
\quad + BiLSTM  & 82.1 & 84.1 & 75.7 & 91.6 & 84.9 \\
     \bottomrule
   \end{tabular}
   \caption{Ablation over the pre-training tasks using the \bertbase architecture. ``No NSP'' is trained without the next sentence prediction task. ``LTR \& No NSP'' is trained as a left-to-right LM without the next sentence prediction, like OpenAI GPT. ``+ BiLSTM'' adds a randomly initialized BiLSTM on top of the ``LTR + No NSP'' model during fine-tuning.
   }
   \label{tab:task_ablation}    
\end{table}

%% file: size_ablation_tab.tex
\begin{table}[b]
\begin{center}
{\small
\begin{tabular}{@{}rrrcccc@{}}
  \toprule
  \multicolumn{3}{c}{Hyperparams}      &      & \multicolumn{3}{c}{Dev Set Accuracy} \\
  \midrule
  \#L & \#H &\#A & LM (ppl) & MNLI-m & MRPC &SST-2               \\
  \midrule
  \
   3 &  768 & 12 & 5.84 & 77.9 & 79.8 & 88.4 \\
   6 &  768 &  3 & 5.24 & 80.6 & 82.2 & 90.7 \\
   6 &  768 & 12 & 4.68 & 81.9 & 84.8 & 91.3 \\
  12 &  768 & 12 & 3.99 & 84.4 & 86.7 & 92.9 \\
  12 & 1024 & 16 & 3.54 & 85.7 & 86.9 & 93.3 \\
  24 & 1024 & 16 & 3.23 & 86.6 & 87.8 & 93.7 \\
\bottomrule
\end{tabular}
} 
\end{center}
\caption{\label{tab:size_ablation} Ablation over BERT model size. \#L = the number of layers; \#H = hidden size; \#A = number of attention heads. ``LM (ppl)'' is the masked LM perplexity of held-out training data.}
\end{table}

%% file: pretrained_embeddings_tab.tex

\begin{table}[t]
\small
\centering
 \begin{tabular}{@{}lcc@{}}
\toprule
System & Dev F1 & Test F1 \\
\midrule
ELMo~\cite{peters-etal:2018:_deep}& 95.7 & 92.2 \\
CVT~\cite{clark2018semi} & - & 92.6 \\
CSE~\cite{akbik2018contextual} & - & {\bf 93.1} \\
\midrule
Fine-tuning approach & & \\
\;\;\;\bertlarge  & 96.6 & 92.8 \\
\;\;\;\bertbase& 96.4 & 92.4 \\
\midrule
Feature-based approach (\bertbase) &  &  \\
\;\;\;Embeddings & 91.0 &- \\
\;\;\;Second-to-Last Hidden   & 95.6&- \\
\;\;\;Last Hidden            & 94.9&- \\
\;\;\;Weighted Sum Last Four Hidden        & 95.9&- \\
\;\;\;Concat Last Four Hidden        & 96.1&- \\
\;\;\;Weighted Sum All 12 Layers        & 95.5&- \\
\bottomrule
\end{tabular}
\caption{CoNLL-2003 Named Entity Recognition results. Hyperparameters were selected using the Dev set. The reported Dev and Test scores are averaged over 5 random restarts using those hyperparameters.
}
\label{tab:ner_results}    
\label{tab:pretrained_embeddings}    
\end{table}

%% file: conclusion.tex
\section{Conclusion}
Recent empirical improvements due to transfer learning with language models have demonstrated that rich, unsupervised pre-training is an integral part of many language understanding systems. In particular, these results enable even low-resource tasks to benefit from deep unidirectional architectures. Our major contribution is further generalizing these findings to deep \emph{bidirectional} architectures, allowing the same pre-trained model to successfully tackle a broad set of NLP tasks.

%% file: appendix_main.tex
\begin{center}
{\bf \large{
    Appendix for ``BERT: Pre-training of Deep Bidirectional Transformers for Language Understanding''}}
\end{center}

We organize the appendix into three sections:
\begin{itemize}
   \item Additional implementation details for BERT are presented in Appendix~\ref{appendix:sec:bert_description};
   
   \item Additional details for our experiments are presented in Appendix~\ref{appendix:sec:exp_details}; and

   \item Additional ablation studies are presented in Appendix~\ref{appendix:sec:more_ablation_studies}.
   
   We present additional ablation studies for BERT including:
   \begin{itemize}
      \item Effect of Number of Training Steps; and
      \item Ablation for Different Masking Procedures.
   \end{itemize}
\end{itemize}
\input{bert_details}

\input{experiments_details}

\input{more_ablation}

%% file: bert_details.tex
\section{Additional Details for BERT}
\label{appendix:sec:bert_description}

\begin{figure*}[ht]
\begin{center}
\includegraphics[width=\textwidth]{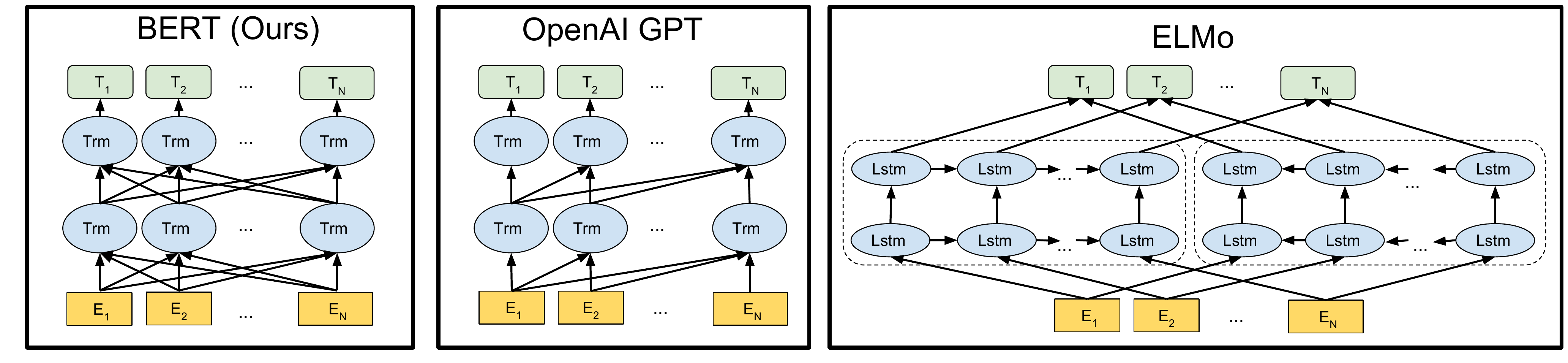}
\end{center}
\caption{Differences in pre-training model architectures. BERT uses a bidirectional Transformer. OpenAI GPT uses a left-to-right Transformer. ELMo uses the concatenation of independently trained left-to-right and right-to-left LSTMs to generate features for downstream tasks. Among the three, only \bert representations are jointly conditioned on both left and right context in all layers. In addition to the architecture differences, BERT and OpenAI GPT are fine-tuning approaches, while ELMo is a feature-based approach.}
\label{fig:BERT_comparisons}
\end{figure*}

\subsection{Illustration of the Pre-training Tasks}
We provide examples of the pre-training tasks in the following.

\paragraph{Masked LM and the Masking Procedure} 

Assuming the unlabeled sentence is {\tt {\small my dog is hairy}}, 
and during the random masking procedure we chose the 4-th token (which corresponding to {\tt {\small hairy}}), our masking procedure
can be further illustrated by
\begin{itemize}
\item 80\% of the time: Replace the word with the {\tt [MASK]} token, e.g., {\tt {\small my dog is hairy $\rightarrow$ my dog is [MASK]}}
\item 10\% of the time: Replace the word with a random word, e.g., {\tt {\small my dog is hairy $\rightarrow$ my dog is apple}}
\item 10\% of the time: Keep the word unchanged, e.g., {\tt {\small my dog is hairy $\rightarrow$ my dog is hairy}}. The purpose of this is to bias the representation towards the actual observed word.
\end{itemize}

The advantage of this procedure is that the Transformer encoder does not know which words it will be asked to predict or which have been replaced by random words, so it is forced to keep a distributional contextual representation of {\it every} input token. Additionally, because random replacement only occurs for 1.5\% of all tokens (i.e., 10\% of 15\%), this does not seem to harm the model's language understanding capability. In Section~\ref{appendix:sec:different_masks},
we evaluate the impact this procedure.

Compared to standard langauge model training, the masked LM only
make predictions on 15\% of tokens in each batch, which suggests that more pre-training steps may be required for the model to converge. In Section~\ref{sec:num_training_steps} we demonstrate that MLM does converge marginally slower than a left-to-right model (which predicts every token), but the empirical improvements of the MLM model far outweigh the increased training cost.

\paragraph{Next Sentence Prediction} 

The next sentence prediction task can be illustrated in the following examples.
\begin{align*}
\text{Input\;} &= \text{\tt {\scriptsize [CLS] the man went to [MASK] store [SEP]}} \\ 
& \text{\tt {\scriptsize \;\;\;\;\;\;\;he bought a gallon [MASK] milk [SEP]}}\\
\text{Label} &= \text{\tt {\scriptsize IsNext}} \\
\\
\text{Input\;} &= \text{\tt {\scriptsize [CLS] the man [MASK] to the store [SEP]}}\\
&\text{\tt {\scriptsize \;\;\;\;\;\;\;penguin [MASK] are flight \#\#less birds [SEP]}}\\
\text{Label} &= \text{\tt {\scriptsize NotNext}}
\end{align*}

\subsection{Pre-training Procedure}
\label{sec:pretraining_procedure}

To generate each training input sequence, we sample two spans of text from the corpus, which we refer to as ``sentences'' even though they are typically much longer than single sentences (but can be shorter also). The first sentence receives the {\tt A} embedding and the second receives the {\tt B} embedding. 50\% of the time {\tt B} is the actual next sentence that follows {\tt A} and 50\% of the time it is a random sentence, which is done for the ``next sentence prediction'' task. They are sampled such that the combined length is $\le$ 512 tokens. The LM masking is applied after WordPiece tokenization with a uniform masking rate of 15\%, and no special consideration given to partial word pieces.

We train with batch size of 256 sequences (256 sequences * 512 tokens = 128,000 tokens/batch) for 1,000,000 steps, which is approximately 40 epochs over the 3.3 billion word corpus. We use Adam with learning rate of 1e-4, ${\beta}_1=0.9$, ${\beta}_2=0.999$, L2 weight decay of $0.01$, learning rate warmup over the first 10,000 steps, and linear decay of the learning rate. We use a dropout probability of 0.1 on all layers. We use a {\tt gelu} activation \cite{hendrycks:2016} rather than the standard {\tt relu}, following OpenAI GPT. The training loss is the sum of the mean masked LM likelihood and the mean next sentence prediction likelihood.

Training of \bertbase was performed on 4 Cloud TPUs in Pod configuration (16 TPU chips total).\footnote{https://cloudplatform.googleblog.com/2018/06/Cloud-TPU-now-offers-preemptible-pricing-and-global-availability.html} Training of \bertlarge was performed on 16 Cloud TPUs (64 TPU chips total). Each pre-training took 4 days to complete.

Longer sequences are disproportionately expensive because attention is quadratic to the sequence length. To speed up pretraing in our experiments, we pre-train the model with sequence length of 128 for 90\% of the steps. Then, we train
the rest 10\% of the steps of sequence of 512 to learn the positional embeddings.

\subsection{Fine-tuning Procedure}
For fine-tuning, most model hyperparameters are the same as in pre-training, with the exception of the batch size, learning rate, and number of training epochs. The dropout probability was always kept at 0.1. The optimal hyperparameter values are task-specific, but we found the following range of possible values to work well across all tasks:

\begin{itemize}[noitemsep]
\item {\bf Batch size}: 16, 32
\item {\bf Learning rate (Adam)}: 5e-5, 3e-5, 2e-5
\item {\bf Number of epochs}: 2, 3, 4
\end{itemize}

We also observed that large data sets (e.g., 100k+ labeled training examples) were far less sensitive to hyperparameter choice than small data sets. Fine-tuning is typically very fast, so it is reasonable to simply run an exhaustive search over the above parameters and choose the model that performs best on the development set.

\subsection{Comparison of BERT, ELMo ,and OpenAI GPT}
\label{appendix:sec:comparing_bert_and_openai}

Here we studies the differences in recent popular representation learning
models including ELMo, OpenAI GPT and BERT.
The comparisons between the model
architectures are shown visually in Figure~\ref{fig:BERT_comparisons}. Note that
in addition to the architecture differences, BERT and OpenAI GPT are fine-tuning approaches, while ELMo is a feature-based approach.

The most comparable existing pre-training method to BERT is OpenAI GPT, which trains a left-to-right Transformer LM on a large text corpus. 
In fact, many of the design decisions in BERT were intentionally made to make it as close to GPT as possible so that the two methods could be minimally compared.
The core argument of this work is that the bi-directionality and the two pre-training tasks presented in Section~\ref{sec:pretraining_tasks} account for the majority of the empirical improvements, but we do note that there are several other differences between how BERT and GPT were trained:

\begin{itemize}
\item GPT is trained on the BooksCorpus (800M words); BERT is trained on the BooksCorpus (800M words) and Wikipedia (2,500M words).
\item GPT uses a sentence separator ({\tt [SEP]}) and classifier token ({\tt [CLS]}) which are only introduced at fine-tuning time; BERT learns {\tt [SEP]}, {\tt [CLS]} and sentence {\tt A}/{\tt B} embeddings during pre-training.
\item GPT was trained for 1M steps with a batch size of 32,000 words; BERT was trained for 1M steps with a batch size of 128,000 words.
\item GPT used the same learning rate of 5e-5 for all fine-tuning experiments; BERT chooses a task-specific fine-tuning learning rate which performs the best on the development set.
\end{itemize}

To isolate the effect of these differences, we perform ablation experiments in Section~\ref{sec:task_ablation} which demonstrate that the majority of the improvements are in fact coming from the two pre-training tasks and the bidirectionality they enable.

\subsection{Illustrations of Fine-tuning on Different Tasks}
\label{appendix:sec:fine_tune_details_and_figures}

\begin{figure*}[ht]
\begin{center}
\includegraphics[width=0.85\textwidth]{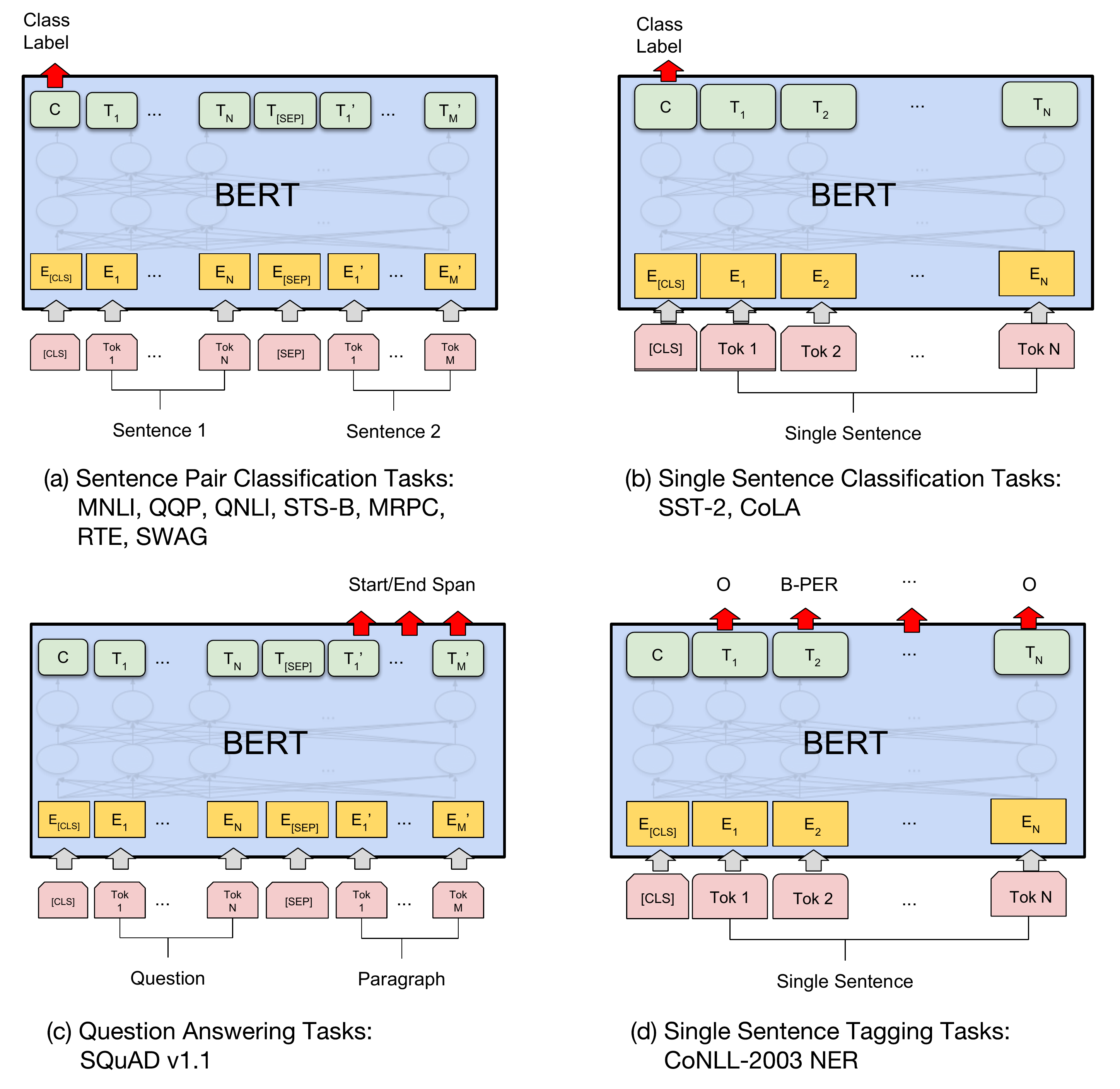}
\end{center}
\caption{Illustrations of Fine-tuning BERT on Different Tasks.}
\label{fig:bert_fine_tune}
\end{figure*}

The illustration of fine-tuning BERT on different tasks can be seen in
Figure~\ref{fig:bert_fine_tune}. Our task-specific models are formed by incorporating \bert with one additional output layer, so a  minimal number of parameters need to be learned from scratch. Among the tasks, (a) and (b) are sequence-level tasks while (c) and (d) are token-level tasks. In the figure, $E$ represents the input embedding, $T_i$ represents the contextual representation of token $i$, \textsc{[CLS]} is the special symbol for classification output, and \textsc{[SEP]} is the special symbol to separate non-consecutive token sequences.

%% file: experiments_details.tex
\section{Detailed Experimental Setup}
\label{appendix:sec:exp_details}

\subsection{Detailed Descriptions for the GLUE Benchmark Experiments.}
\label{appendix:sec:glue}

Our GLUE results in Table\ref{tab:glue_official} are obtained from  \url{https://gluebenchmark.com/leaderboard} and \url{https://blog.openai.com/language-unsupervised}.
The GLUE benchmark includes the following datasets, the descriptions of which were originally summarized in \citet{wang-etal:2018:_glue}:

\paragraph{MNLI} Multi-Genre Natural Language Inference is a large-scale, crowdsourced entailment classification task~\cite{williams-nangia-bowman:2018}. Given a pair of sentences, the goal is to predict whether the second sentence is an {\it entailment}, {\it contradiction}, or {\it neutral} with respect to the first one.

\paragraph{QQP} Quora Question Pairs is a binary classification task where the goal is to determine if two questions asked on Quora are semantically equivalent~\cite{chen-etal:2018:_quora}.

\paragraph{QNLI} Question Natural Language Inference is a version of the Stanford Question Answering Dataset~\cite{rajpurkar-etal:2016:_squad} which has been converted to a binary classification task~\cite{wang-etal:2018:_glue}. The positive examples are (question, sentence) pairs which do contain the correct answer, and the negative examples are (question, sentence) from the same paragraph which do not contain the answer.

\paragraph{SST-2} The Stanford Sentiment Treebank is a binary single-sentence classification task consisting of sentences extracted from movie reviews with human annotations of their sentiment~\cite{socher-etal:2013:_recur}.

\paragraph{CoLA} The Corpus of Linguistic Acceptability is a binary single-sentence classification task, where the goal is to predict whether an English sentence is linguistically ``acceptable'' or not~\cite{warstadt-singh-bowman:2018:_corpus}.

\paragraph{STS-B} The Semantic Textual Similarity Benchmark is a collection of sentence pairs drawn from news headlines and other sources~\cite{cer-etal:2017}. They were annotated with a score from 1 to 5 denoting how similar the two sentences are in terms of semantic meaning. 

\paragraph{MRPC} Microsoft Research Paraphrase Corpus consists of sentence pairs automatically extracted from online news sources, with human annotations for whether the sentences in the pair are semantically equivalent~\cite{dolan-brockett:2005:_autom}.

\paragraph{RTE} Recognizing Textual Entailment is a binary entailment task similar to MNLI, but with much less training data~\cite{bentivogli-etal:2009}.\footnote{Note that we only report single-task fine-tuning results in this paper. A multitask fine-tuning approach could potentially push
the performance even further. For example, we did observe substantial improvements on RTE from multi-task training with MNLI.}

\paragraph{WNLI} Winograd NLI is a small natural language inference dataset \cite{levesque-davis-morgenstern:2011:_winog}. The GLUE webpage notes that there are issues with the construction of this dataset,~\footnote{\url{https://gluebenchmark.com/faq}} and every trained system that's been submitted to GLUE has performed worse than the 65.1 baseline accuracy of predicting the majority class. We therefore exclude this set to be fair to OpenAI GPT. For our GLUE submission, we always predicted the majority class.

%% file: more_ablation.tex

\section{Additional Ablation Studies}
\label{appendix:sec:more_ablation_studies}

\subsection{Effect of Number of Training Steps}
\label{sec:num_training_steps}

Figure~\ref{fig:step_abalation} presents MNLI Dev accuracy after fine-tuning from a checkpoint that has been pre-trained for $k$ steps. This allows us to answer the following questions:

\input{steps_ablation_fig}

\begin{enumerate}
\item Question: Does BERT really need such a large amount of pre-training (128,000 words/batch * 1,000,000 steps) to achieve high fine-tuning accuracy? \\
Answer: Yes, \bertbase achieves almost 1.0\% additional accuracy on MNLI when trained on 1M steps compared to 500k steps.
\item Question: Does MLM pre-training converge slower than LTR pre-training, since only 15\% of words are predicted in each batch rather than every word? \\
Answer: The MLM model does converge slightly slower than the LTR model. However, in terms of absolute accuracy the MLM model begins to outperform the LTR model almost immediately.
\end{enumerate}

\subsection{Ablation for Different Masking Procedures}
\label{appendix:sec:different_masks}

In Section~\ref{sec:pretraining_tasks}, we mention that BERT uses a mixed strategy for masking the target tokens when pre-training with the masked language
model (MLM) objective. The following is an ablation study to evaluate the effect of different masking strategies.

Note that the purpose of the masking strategies is to reduce the mismatch
between pre-training and fine-tuning, as the {\tt [MASK]} symbol never appears during the fine-tuning stage. We report the Dev results for both
MNLI and NER. For NER, we report both fine-tuning and feature-based approaches,
as we expect the mismatch will be amplified for the feature-based approach as
the model will not have the chance to adjust the representations.

\input{mask_procedure_ablation_tab}

The results are presented in Table~\ref{tab:mask_ablation}. In the table,
 \textsc{Mask} means that we replace the target token with the {\tt [MASK]} symbol
for MLM; \textsc{Same} means that we keep the target token as is; \textsc{Rnd}
means that we replace the target token with another random token. 

The numbers in the left part of the table represent the probabilities of the specific strategies used during MLM pre-training (BERT uses 80\%, 10\%, 10\%). The right part of the paper represents the Dev set results. For the feature-based
approach, we concatenate the last 4 layers of BERT as the features, which
was shown to be the best approach in Section~\ref{sec:ner}.

From the table it can be seen that fine-tuning is surprisingly robust to
different masking strategies. However, as expected, using only the \textsc{Mask} strategy
was problematic when applying the feature-based approach to NER. Interestingly,
using only the \textsc{Rnd} strategy performs much worse than our strategy as well.

%% file: steps_ablation_fig.tex

\begin{figure}[b]
\begin{tikzpicture}
 \begin{axis}[
   width=0.95\columnwidth,
   height=0.7\columnwidth,
   legend cell align=left,
   legend style={at={(1, 0)},anchor=south east,font=\scriptsize},
   mark options={mark size=3},
   font=\scriptsize,
   xmin=0, xmax=1000,
   ymin=75, ymax=85,
   xtick={200,400,600,800,1000},
   ymajorgrids=true,
   xmajorgrids=true,
   xlabel style={yshift=0.5ex,},
   xlabel=Pre-training Steps (Thousands) ,
   ylabel=MNLI Dev Accuracy,
   ylabel style={yshift=-0.5ex,}]
    \addplot[mark=triangle,g-blue] plot coordinates {
      (30, 78.6)
      (50, 79.6)
      (100, 80.5)
      (200, 82.2)
      (400, 83.2)
      (600, 84.0)
      (800, 84.3)
      (1000, 84.4)
    };
    \addlegendentry{\bertbase (Masked LM)}
    \addplot[mark=x,g-red] plot coordinates {
      (30, 79.4)
      (50, 79.8)
      (100, 80.3)
      (200, 81.0)
      (400, 81.7)
      (600, 81.9)
      (800, 82.1)
      (1000, 82.2)
    };
    \addlegendentry{\bertbase (Left-to-Right)}
     \end{axis}
\end{tikzpicture}
\caption{Ablation over number of training steps. This shows the MNLI accuracy after fine-tuning, starting from model parameters that have been pre-trained for $k$ steps. The x-axis is the value of $k$.}
\label{fig:step_abalation}
\end{figure}

%% file: mask_procedure_ablation_tab.tex
\begin{table}[ht]
\begin{center}
{\small
\begin{tabular}{@{}rrrccc@{}}
  \toprule
  \multicolumn{3}{c}{Masking Rates} & \multicolumn{3}{c}{Dev Set Results} \\
  \cmidrule(r{0.2cm}){1-3}
  \cmidrule(l{0.2cm}){4-6}
  \textsc{Mask} &	\textsc{Same}&	\textsc{Rnd}	& {MNLI} &	\multicolumn{2}{c}{NER}\\
   &  & & {\footnotesize Fine-tune} &   {\footnotesize Fine-tune}	& {\footnotesize Feature-based}\\
    \cmidrule(r{0.2cm}){1-3}
  \cmidrule(l{0.1cm}r{0.1cm}){4-4}
  \cmidrule(l{0.2cm}){5-6}
  
  80\%	&10\%	&10\%	&84.2	&95.4	&94.9\\
100\%	&0\%	&0\%	&84.3	&94.9	&94.0\\
80\%	&0\%	&20\%	&84.1	&95.2	&94.6\\
80\%	&20\%	&0\%	&84.4	&95.2	&94.7\\
0\%	&20\%	&80\%	&83.7	&94.8	&94.6 \\
0\%	&0\%	&100\%	&83.6	&94.9	&94.6\\
\bottomrule
\end{tabular}
} 
\end{center}
\caption{\label{tab:mask_ablation} Ablation over different masking strategies.}
\end{table}